\documentclass[10pt,twocolumn,letterpaper]{article}

\usepackage[pagenumbers]{cvpr} 

\definecolor{cvprblue}{rgb}{0.21,0.49,0.74}
\usepackage[pagebackref,breaklinks,colorlinks,allcolors=cvprblue]{hyperref}
\usepackage[table]{xcolor}
\usepackage{multirow}
\usepackage{algorithm}
\usepackage{algpseudocode}
\usepackage{array} 

\def\paperID{36474} 
\def\confName{CVPR}
\def\confYear{2026}

\title{Think-as-You-See: Streaming Chain-of-Thought Reasoning for Large Vision-Language Models}

\author{
Jialiang Zhang$^{1,2}$\thanks{Equal contribution. \quad \textdagger Corresponding Author.} \quad
Junlong Tong$^{1,3}$\footnotemark[1] \quad
Junyan Lin$^{1,4}$\footnotemark[1] \quad
Hao Wu$^{1}$ \quad \\
Yirong Sun$^{1}$ \quad
Yunpu Ma$^{5}$ \quad
Xiaoyu Shen$^{1,6}$\textdagger\\
$^{1}$Institute of Digital Twin, Eastern Institute of Technology, Ningbo
$^{2}$Ocean University of China\\
$^{3}$Shanghai Jiao Tong University
$^{4}$The Hong Kong Polytechnic University\\
$^{5}$Munich Center for Machine Learning, LMU Munich\\
$^{6}$Ningbo Key Laboratory of Spatial Intelligence and Digital Derivative\\
{\tt\small zhangjia\_liang@foxmail.com \quad xyshen@eitech.edu.cn}
}

\begin{document}
\maketitle
\begin{abstract}
Large Vision-Language Models (LVLMs) have made significant strides in video reasoning, yet most existing systems rely on a batch inference paradigm that processes the entire video before reasoning begins. This ``wait-and-see'' approach neglects the inherently streaming nature of real-world video, introducing substantial latency and exacerbating temporal drift. In this paper, we propose Think-as-You-See (TaYS), a framework that shifts LVLMs toward a streaming reasoning paradigm, enabling continuous, incremental inference synchronized with the visual stream. We introduce three key innovations: (1) a streaming attention mask to enforce temporal causality; (2) a decoupled positional encoding strategy to resolve cross-modal index conflicts; and (3) a parallel dual KV-cache mechanism that decouples visual encoding from reasoning generation, enabling concurrent frame ingestion and token decoding. Empirical evaluations on the VideoEspresso benchmark using the Qwen2.5-VL family demonstrate that TaYS improves reasoning accuracy by 2.9\%, reduces Time-to-First-Token (TTFT) from 10.6s to near-zero, and cuts reasoning-event deviation by 55\%. Our results suggest that aligning LVLM reasoning with the streaming nature of video is a vital step toward responsive, real-time multimodal intelligence. The code is available at \href{https://github.com/EIT-NLP/StreamingLLM/tree/main/TaYS}{this repository.}
\end{abstract}    
\begin{figure}[t]
  \centering
   \includegraphics[width=\linewidth]{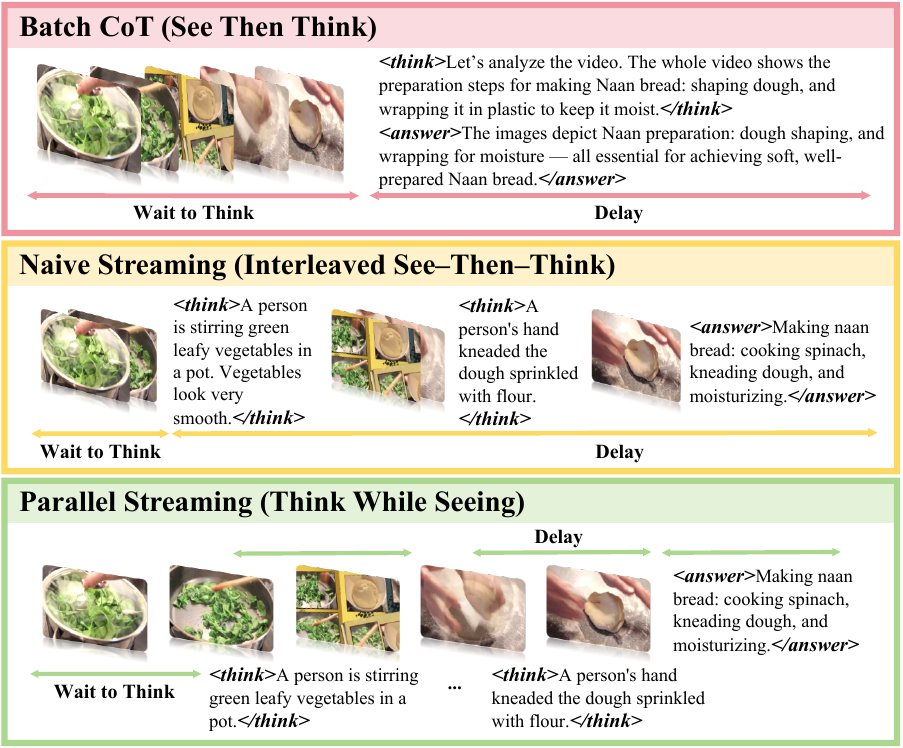}
   \caption{Conventional LVLM reasoning adheres to the \textit{batch thinking} paradigm, deferring inference until the entire input is received. This approach often leads to high latency and uneven attention allocation across inputs. In contrast, our proposed \textit{streaming thinking} paradigm enables LVLMs to reason concurrently with input reception, thereby reducing latency and ensuring consistency between attention and input order.}
   \label{fig:figure1}
\end{figure}

\section{Introduction}

Large Vision-Language Models (LVLMs) have recently achieved remarkable milestones in multimodal reasoning~\cite{yin2024survey,li2025surveystateartlarge}, as demonstrated by state-of-the-art systems such as GPT-4o~\cite{hurst2024gpt}, Gemini~\cite{comanici2025gemini} and Qwen-VL~\cite{bai2025qwen3vltechnicalreport}. Despite these advancements, a pervasive bottleneck remains: the vast majority of LVLM-based video reasoning systems are anchored to a \emph{batch inference} paradigm where the model requires the full video to be available offline before processing begins~\cite{chen2024videollmonlineonlinevideolarge,rao2025universalsoccervideounderstanding}. Under this ``wait-and-see'' paradigm, both the information density and the computational complexity scale directly with video length, making accurate and coherent interpretation increasingly difficult~\cite{lin2024video,weng2024longvlmefficientlongvideo,huang2025prunevid,Wu2026survey,fan2026what}.

Current research attempts to mitigate this issue using Chain-of-Thought (CoT) reasoning~\cite{himakunthala-etal-2023-lets, pmlr-v235-fei24a, 10.1145/3746027.3755837, zheng2023ddcot, ghazanfari2025chain} paired with auxiliary modules for explicit frame referencing~\cite{arnab2025temporal, Han_2025_CVPR, ge2025famemind, zhang2023multimodal}. By grounding predictions in specific keyframes and reasoning traces, these methods enhance both interpretability and accuracy. However, they are still restricted to the same \emph{batch} inference paradigm. As the temporal window of the input video expands, the delay between a visual event and the model's corresponding reasoning step grows proportionally~\cite{huang2025revisitingmultimodalpositionalencoding,luo2025thinking}. This latency accumulation often leads to ``temporal drift'', where the model loses track of early cues, resulting in significant hallucinations and a loss of contextual coherence~\cite{wang2024videohallucer, zhang2024eventhallusion, chen2025towards}.

This batch-processing assumption is increasingly at odds with the demands of the real world. In domains such as robotics teleoperation, autonomous driving, and live surveillance, video is not a static file but an \emph{evolving stream}~\cite{tong2026staticinferencedynamicinteraction}. Human cognition naturally does not wait for a sequence to end before processing; rather, we update our mental models incrementally as new evidence unfolds~\cite{graesser1994constructing,stenning2012human}. Bridging this gap requires a paradigm shift: models must transition from post-hoc analysis to active, concurrent understanding~\cite{tong2025streamingthinker}.

Motivated by this streaming characteristics of video, we propose \textbf{Think-as-You-See (TaYS)}, a unified framework that equips LVLMs with streaming video CoT capabilities.  
In this framework, reasoning is not a terminal step but a continuous process that evolves in tandem with the visual stream. This approach ensures that inference trajectories are progressively refined, minimizing cognitive lag and ensuring that reasoning is always synchronized with the most relevant visual context. 

A naive implementation that supports this framework is interleaved streaming where the model alternatingly processes a video segment and generates a corresponding reasoning trace~\cite{tong2026staticinferencedynamicinteraction}. This implementation, however, is fundamentally limited by its sequential nature. This ``blocking'' mechanism forces the model to pause visual ingestion until token generation is complete, creating a computational bottleneck that contradicts the fluid nature of live video this implementation~\cite{tong2025llm,lin2026speak}. 
To overcome this, TaYS harmonizes stream-aligned training with true parallel inference via three key innovations: \emph{(1) a streaming attention mask} to enforce temporal causality, \emph{(2) a decoupled positional encoding strategy} that independently indexes visual and reasoning tokens to avoid cross-modal index conflicts, and \emph{(3) a parallel dual KV-cache mechanism} that decouples visual encoding from reasoning generation, enabling concurrent frame ingestion and token decoding.

We instantiate TaYS on the Qwen2.5-VL family~\cite{Qwen2.5-VL} and evaluate its efficacy across tasks requiring complex event dynamics and causal reasoning. On the extended VideoEspresso~\cite{Han_2025_CVPR} benchmark, TaYS improves reasoning accuracy by \emph{+2.9\%} over batch CoT baselines and achieves a \emph{43.7\% win rate} in human-aligned GPT-5 evaluations. Critically, TaYS reduces the Time-to-First-Token (TTFT) from \emph{10.6s} in batch mode to nearly \emph{zero}, while improving temporal grounding by reducing reasoning-event deviation from \emph{1.52s} to \emph{0.69s}. These results demonstrate that aligning LVLM reasoning with the streaming nature of video is not only biologically intuitive but also a practical necessity for the next generation of real-time AI applications.

\paragraph{Contributions.}
Our contributions are fourfold:
\begin{itemize}
    \item We introduce a principled streaming reasoning paradigm for LVLMs, enabling incremental, temporally grounded inference aligned with unfolding visual evidence.
    
    \item We design a cohesive training and inference architecture that operationalizes streaming reasoning, combining causal masking, decoupled positional encoding, and a parallel dual-cache mechanism.
    
    \item We conduct comprehensive empirical evaluations on streaming video reasoning tasks, demonstrating improved reasoning quality and significantly enhanced responsiveness compared to batch and interleaved baselines.
\end{itemize}

\section{Related Work}

\paragraph{Multimodal Chain-of-Thought Reasoning.}
Multimodal reasoning enables LVLMs to integrate visual and textual information for complex decision making. 
Existing approaches generally fall into two paradigms. 
The first, \emph{text-centric reasoning}, converts visual inputs into captions or symbolic descriptions, enabling subsequent linguistic inference~\cite{hu2022promptcap, zheng2023ddcot, wang-etal-2024-videocot, ghazanfari2025chain, hu2025streamingcot, zhang2023multimodal}. 
While effective for interpretability, this pipeline assumes full input availability before reasoning, leading to high latency and weak temporal grounding~\cite{luo2025thinking}. 

The second paradigm, \emph{interleaved multimodal reasoning}, alternates visual and textual tokens to promote more structured cross-modal interaction~\cite{himakunthala-etal-2023-lets, Han_2025_CVPR, 10.1145/3746027.3755837, arnab2025temporal, ge2025famemind, pmlr-v235-fei24a, su2025thinking, cheng2025comt}. 
Although this improves transparency and causal interpretability, it typically relies on sequential processing and explicit intermediate generation, which increases inference latency and computational overhead. 

Recent works also explore efficiency-oriented designs, such as adaptive reasoning depth~\cite{lu2025prolonged,fan2025visipruner,ding2026llms,wu2026hidrop,liu2026vica}, compact cot tokens~\cite{xiang2025can, peng2025skywork,shen2025efficient,zhao2026policy}. 
However, these studies primarily optimize computation under offline settings, and do not explicitly address temporally grounded, low-latency reasoning over streaming inputs.

\paragraph{Streaming and Memory-Based Video Understanding.}
The demand for real-time multimodal systems has stimulated research on \emph{streaming video understanding}, where models process frames incrementally instead of in batch mode~\cite{chen2024videollm, qian2024streaming, tong2025streamingthinker, lin2026speak}. 
Representative efforts focus on streaming captioning, multi-round QA, and conversational agents~\cite{chen2025livecc, qian2025dispider, xiong2025streamingvideounderstandingmultiround, di2025rekv, Chatterjee_2025_ICCV, xu2025streamingvlm, liu2024streamchat, yang2025streamagent}. 
While these approaches improve temporal consistency and enable online interaction, they often emphasize description or response continuity rather than explicit, stepwise reasoning aligned with evolving visual evidence.

Another line of work leverages memory mechanisms or temporal compression to maintain long-context representations efficiently. \cite{he2024malmm, 10.5555/3692070.3692171, shen2024longvu}
By aggregating or consolidating historical features, these methods reduce computational cost but may sacrifice fine-grained temporal alignment and incremental interpretability. \cite{lee2024video, NEURIPS2024_d7ce06e9}
In contrast, our formulation does not compress or abstract away temporal structure; instead, it explicitly synchronizes reasoning generation with frame-level updates through causal masking, decoupled positional encoding, and parallel cache management.

Overall, existing works either assume offline reasoning or prioritize temporal summarization over progressive inference. 
Our TaYS framework complements these directions by focusing on \emph{true streaming reasoning}, where perception and reasoning evolve concurrently under strict temporal causality, enabling low-latency and temporally grounded video understanding.
\section{Methodology}
This section presents TaYS, a supervised fine-tuning framework that integrates streaming video CoT generation with streaming training and inference mechanisms. Its objective is to adapt batch-oriented Large Vision-Language Models to the streaming thinking paradigm.

\subsection{Task Definition and Preliminaries}
Streaming Video CoT demands that a model continuously process a video stream, performing temporal reasoning on queries regarding previously observed visual content at arbitrary time steps. In this section, we formalize this task and highlight its fundamental distinctions from the conventional offline paradigm.

\paragraph{Streaming Video CoT vs. Offline Video CoT.}
Formally, let a video stream be represented as a sequence of visual frames $\mathcal{V} = \{F_t \mid 1 \le t \le T\}$, and let $C_{<t}$ denote the accumulated multimodal context prior to time $t$ (e.g., historical textual or visual reasoning states).

\textbf{Offline Video CoT.}
In the offline setting, the model assumes global access to all frames in $\mathcal{V}$ before generating any reasoning tokens.
At the final time step $t = T$, the reasoning process is formulated as:
\begin{equation}
\begin{aligned}
h_i &= \mathrm{Decoder}\big(y_{<i}; \, \mathrm{Enc}(\mathcal{V})\big), \\
\hat{y}_i &\sim P_\theta(y_i \mid \mathcal{V}, y_{<i}),
\end{aligned}
\label{eq:offline_vcot}
\end{equation}
where $\mathrm{Enc}(\mathcal{V})$ encodes the complete frame sequence $\{F_1, \ldots, F_T\}$, and $y_i$ denotes the $i$-th reasoning token.
Consequently, Offline Video CoT optimizes the joint probability over the entire sequence:
\begin{equation}
\max_\theta \; P_\theta(Y \mid \mathcal{V})
= \prod_{i=1}^{N} P_\theta(y_i \mid \mathcal{V}, y_{<i}),
\end{equation}
which necessitates full video observation prior to the onset of generation.

\textbf{Streaming Video CoT.}
Conversely, Streaming Video CoT performs \textit{incremental reasoning} as frames arrive.
At any time step $t$, only the partial frame sequence $\mathcal{V}_{\le t} = \{F_1, \ldots, F_t\}$ is observable. The model generates reasoning tokens conditioned on this partial visual context and the prior reasoning states:
\begin{equation}
\begin{aligned}
h_i^t &= \mathrm{Decoder}\big(y_{<i}^t; \, \mathrm{Enc}(\mathcal{V}_{\le t}), C_{<t}\big), \\
\hat{y}_i^t &\sim P_\theta(y_i^t \mid \mathcal{V}_{\le t}, y_{<i}^t, C_{<t}).
\end{aligned}
\label{eq:stream_vcot}
\end{equation}
In contrast to Eq.~\ref{eq:offline_vcot}, the model is prohibited from accessing unseen future frames $\{F_{t+1}, \ldots, F_T\}$, enforcing a strict causal constraint on both visual and linguistic modalities.
This paradigm optimizes the cumulative probability up to time $t$:
\begin{equation}
\max_\theta \; P_\theta(Y_{\le t} \mid \mathcal{V}_{\le t})
= \prod_{i=1}^{N_t} P_\theta(y_i^t \mid \mathcal{V}_{\le t}, y_{<i}^t, C_{<t}),
\end{equation}
where $N_t$ denotes the number of reasoning tokens generated up to time $t$.

Architecturally, Streaming Video CoT updates its reasoning states concurrently with incoming frames, whereas Offline Video CoT encodes the entire video before reasoning commences. Notably, Offline Video CoT can be viewed as a degenerate case of Streaming Video CoT, wherein all reasoning is deferred until the video stream terminates.

\paragraph{Design Principles.}
To facilitate real-time reasoning, Streaming Video CoT leverages the causal structure of LLM decoders to balance efficiency and accuracy while minimizing redundant computation. 
During streaming, KV-Caches are incrementally stored and reused as contextual memory, enabling state updates without re-encoding historical frames. 
A causal attention mask restricts token access to future information, ensuring that each video token attends exclusively to past visual inputs and prior reasoning states. 
This architecture effectively disentangles temporal visual processing from linguistic reasoning, achieving efficient and temporally consistent inference across dynamic video streams.

\begin{figure}[t]
  \centering
   \includegraphics[width=\linewidth]{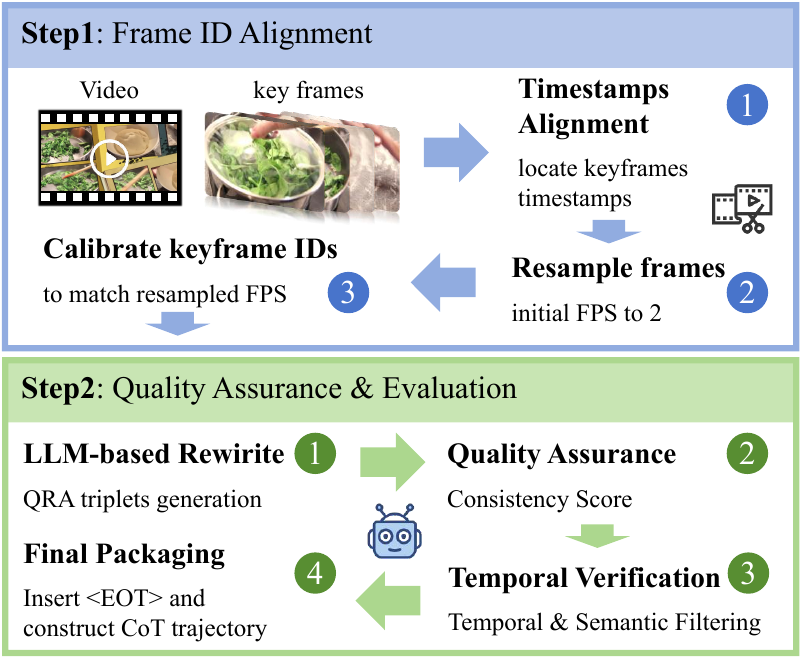}
   \caption{Overview of the two-step process for generating Streaming Video CoT. \textbf{Step 1} Adjust the frame ID while maintaining frame caption alignment. \textbf{Step 2} Generate a progressive frame aware trajectory using the original annotations.}
   \label{fig:figure2}
\end{figure}

\subsection{Streaming Video CoT Generation}

To enable temporally grounded incremental reasoning, we construct a streaming-style Video CoT dataset that departs from conventional batch reasoning trajectories, which assume full-video access and overlook progressive reasoning behavior. Our construction is based on the training split of \textsc{VideoEspresso}, which contains temporally coherent videos annotated with keyframe-level descriptions capturing causal and logical transitions. These keyframes serve as semantic anchors for extracting frame-aligned reasoning trajectories under streaming constraints. The overall pipeline is illustrated in Figure~\ref{fig:figure2}, with additional details provided in Appendix~\ref{sec:details_cot}.

\paragraph{Frame ID Alignment.}
To ensure strict temporal alignment between visual inputs and reasoning units, we adopt timestamp-based resampling instead of uniform frame sampling. All videos are resampled to 2 FPS. For each target sampling timestamp $\tau'_{t'} = 0.5(t'-1)$ seconds, the selected frame $F_{t'}$ is defined as:

\begin{equation} F_{t'} \! = \begin{cases} \! F_k, \text{if } \tau'_{t'} \in [\tau_k^{\text{start}}, \tau_k^{\text{end}}] \& F_k \text{ is a keyframe},\\ \! \arg\min_{F_t} |\tau_t - 0.5(t'-1)|, \text{otherwise.} \end{cases} \label{eq:resample} \end{equation}

where $\{\tau_t\}_{t=1}^{T}$ denote original frame timestamps.  
This strategy preserves annotated moments while maintaining temporal regularity. After resampling, frame indices are re-normalized and clips are truncated to the model’s maximum input length, ensuring consistency among visual frames, timestamps, and textual annotations.

\paragraph{Structured Trajectory Construction.}
Each aligned keyframe $F_t$ is associated with a reasoning sentence $R_t$ and visual evidence $E_t$. To construct structured reasoning trajectories, we prompt GPT-4o~\cite{hurst2024gpt} to generate triplets $(Q_t, R_t, A_t)$ representing the temporally grounded question, reasoning step, and answer derived from the annotated content. This enforces frame-level incremental reasoning and yields temporally segmented reasoning units across the video.

\paragraph{Quality Control.}
To ensure semantic coherence and temporal consistency, we compute an alignment score between each question and its corresponding reasoning sentence:

\begin{equation}
\mathrm{consistency}(Q_t, R_t)
=
\frac{v_Q \cdot v_R}{\|v_Q\| \, \|v_R\|},
\end{equation}

where $v_Q$ and $v_R$ are embedding vectors obtained from the BGE-M3 model~\cite{chen-etal-2024-m3}. Samples with low semantic alignment or temporal inconsistency are discarded. The remaining instances form high-quality streaming reasoning trajectories.

Finally, sentence-level boundary tokens \texttt{<EOT>} are inserted to delimit minimal reasoning units, encouraging the model to generate causally ordered and frame-consistent outputs conditioned only on preceding visual observations.

\begin{figure*}[t]
  \centering
  \includegraphics[width=\linewidth]{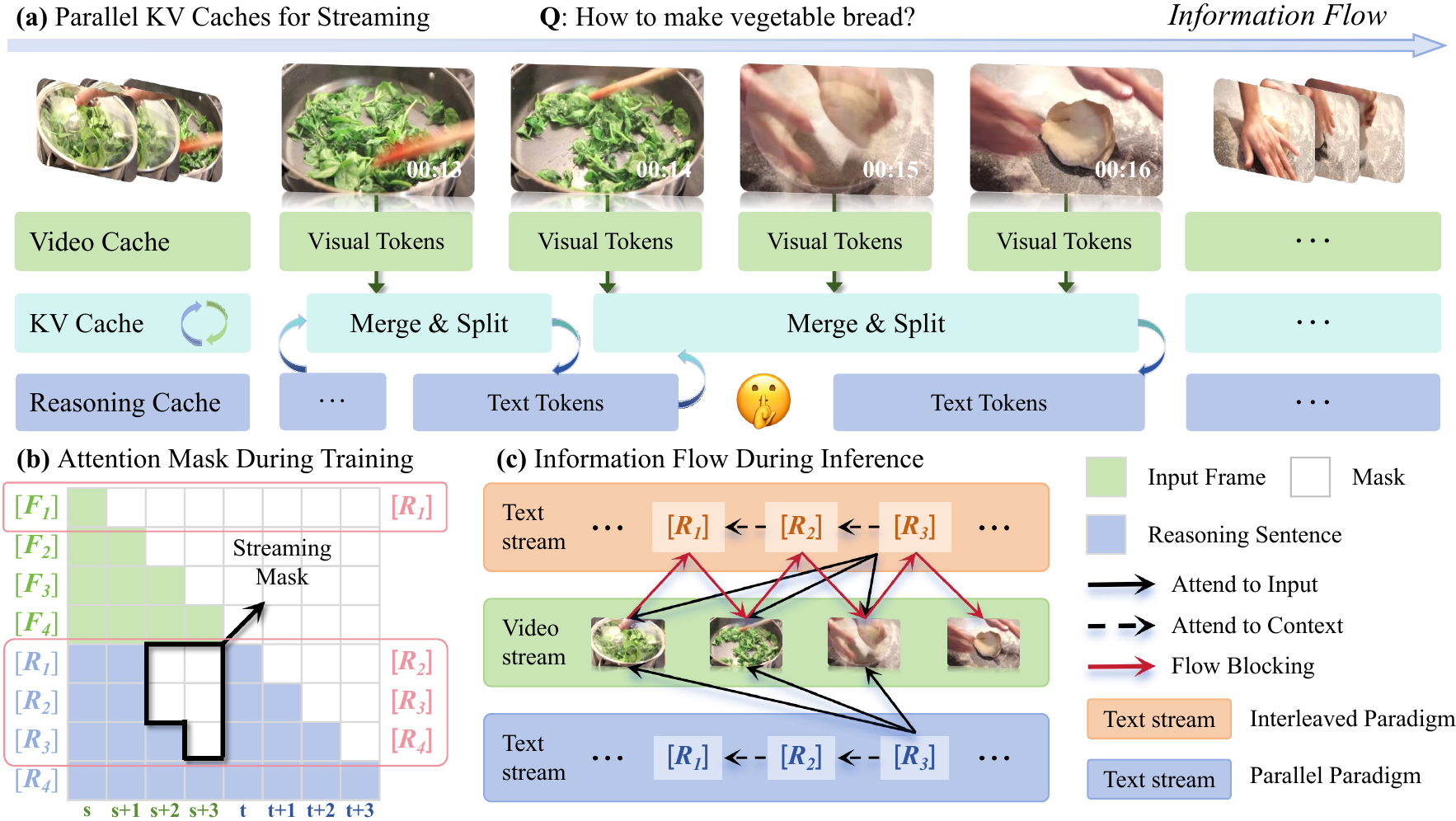}
  \caption{Overview of the streaming reasoning framework. \textbf{(a)}~Parallel video reasoning KV caches enable concurrent visual encoding and reasoning generation via dynamic merge and split operations. \textbf{(b)}~The streaming attention mask enforces causal alignment between frames and reasoning steps. \textbf{(c)}~During inference, parallel information flow reduces attention path length and alleviates sequential blocking compared with interleaved paradigms.}
  \label{fig:main_figure}
\end{figure*}

\subsection{Naive Streaming Paradigm}

A straightforward way to emulate streaming behavior is to interleave video and reasoning tokens during training. 
Concretely, each frame $F_t$ is immediately followed by its associated reasoning segment $R_t$, forming an alternating sequence 
$\{F_1, R_1, F_2, R_2, \dots, F_T, R_T\}$.
All visual and textual embeddings are concatenated into a single causal token stream and processed autoregressively.

This strict interleaving imposes a serialized dependency between perception and reasoning.
Since all tokens share a single causal attention space, new visual tokens cannot be encoded until the preceding reasoning tokens are generated, and reasoning cannot proceed until visual tokens are appended.
Such coupling creates a computational bottleneck and prevents concurrent updates across modalities.

Although this design superficially resembles a “thinking-while-watching” process, it tightly entangles perception and reasoning in a way that deviates from the pretraining distribution of LVLMs, where visual encoding and textual decoding are typically factorized.
As illustrated in Figure~\ref{fig:main_figure}(c), this paradigm therefore suffers from reduced efficiency and limited scalability in long streaming scenarios.

\subsection{Parallel Streaming Paradigm}

To overcome the intrinsic serialization bottleneck of naive interleaving strategies, we introduce a parallel streaming paradigm termed \emph{Think-as-You-See (TaYS)}. 
Unlike conventional approaches that treat reasoning as a post-hoc process dependent on complete visual encoding, TaYS decouples perception from reasoning while strictly preserving temporal causality. 
This architecture enables concurrent execution of visual ingestion and cognitive inference, bridging the gap between streaming perception and real-time reasoning.

\paragraph{Streaming Attention Mask.}

In streaming scenarios, maintaining strict temporal causality is paramount: a reasoning step at time $t$ must strictly attend to visual evidence accumulated up to $t$, remaining agnostic to future frames. 
Standard batch attention mechanisms, which globally expose all visual tokens, violate this causal constraint and are unsuitable for streaming inference.

To address this, we design a streaming-aware attention mask that enforces fine-grained visibility constraints. 
Consider a visual sequence of length $N_v$ and a reasoning sequence of length $N_r$. 
For a query token at position $i$ and a key token at position $j$, the masked attention matrix $\widetilde{M}(i,j)$ is formulated as:

\begin{equation*}
\widetilde{M}(i,j)=
\begin{cases}
-\infty, & i > N_v,\; j < N_v,\; j > i - N_v, \\
M_{\text{causal}}(i,j), & \text{otherwise},
\end{cases}
\end{equation*}

where $M_{\text{causal}}$ represents the standard autoregressive mask. 
The condition $j > i - N_v$ effectively creates a sliding window over the visual tokens relative to the current reasoning step. 
This construction ensures that each reasoning token only integrates information from the current temporal window, preventing information leakage from future frames and ensuring the generated reasoning remains grounded in observed reality.

\paragraph{Streaming Positional Encoding.}

While masking enforces logical visibility, positional encoding must resolve index conflicts arising from the concurrent growth of visual and reasoning streams. 
Modern Large Vision-Language Models (LVLMs) typically employ Rotary Position Embeddings (RoPE)~\cite{su2024roformer}, where relative positional information is encoded via rotation matrices. 
Under standard monolithic indexing, the attention interaction between reasoning token $r_t$ and visual token $v_s$ is computed as:

\begin{equation}
(\mathbf{\mathcal{R}}_{N_v+t}\mathbf{q}_{r_t})^\top
(\mathbf{\mathcal{R}}_{s}\mathbf{k}_{v_s})
=
\mathbf{q}_{r_t}^\top
\mathbf{\mathcal{R}}_{(N_v+t)-s}^\top
\mathbf{k}_{v_s}.
\end{equation}

In this setup, the reasoning position is offset by the total visual length $N_v$. 
However, in a streaming context where $N_v$ expands continuously, this indexing introduces dynamic shifts in relative positions, potentially destabilizing the model's temporal perception. 
To eliminate this interference, we propose a modality-decoupled positional indexing scheme:

\[
\mathrm{pos}(v_s)=s, 
\qquad
\mathrm{pos}(r_t)=t.
\]

This assigns independent positional axes for vision and reasoning. 
The resulting attention mechanism becomes:

\begin{equation}
(\mathbf{\mathcal{R}}_{t}\mathbf{q}_{r_t})^\top
(\mathbf{\mathcal{R}}_{s}\mathbf{k}_{v_s})
=
\mathbf{q}_{r_t}^\top
\mathbf{\mathcal{R}}_{t-s}^\top
\mathbf{k}_{v_s}.
\end{equation}

By isolating the positional spaces, this decoupling prevents index collision and ensures that the relative temporal distance ($t-s$) remains semantically consistent, preserving stable alignment between reasoning updates and visual observations regardless of the growing sequence length.

\paragraph{Attention Pathways.}

The architectural choices in different paradigms fundamentally reshape the information flow. 
Batch reasoning necessitates encoding the entire video prior to decoding, resulting in a long sequential attention path and high initial latency. 
Interleaved reasoning alternates between frame input and text generation but relies on a monolithic cache, creating a sequential dependency that forces the reasoning process to stall during visual encoding.
In contrast, TaYS restructures the dataflow by separating modality-specific memory pathways while enabling dynamic fusion during decoding. 
This design substantially shortens the effective attention path, allowing the model to initiate reasoning immediately upon receiving the first frame without waiting for subsequent visual inputs (Figure~\ref{fig:main_figure}(c)).

\begin{table*}[t]
\centering
\small
\setlength{\tabcolsep}{3pt} 
\caption{Comparison of reasoning accuracy on the extended \textsc{VideoEspresso} benchmark. TaYS consistently achieves competitive or superior performance while maintaining low latency, demonstrating the effectiveness of the streaming reasoning paradigm. In the table, \textbf{bold} numbers denote the best results, and \underline{underlined} numbers indicate the second-best results for each task category.}
\label{tab:accuracy}
\resizebox{\textwidth}{!}{
\begin{tabular}{l|cccccccccccccc|c}
\toprule
\textbf{Model} & \textbf{Narr.} & \textbf{Event} & \textbf{Ingr.} & \textbf{Caus.} & \textbf{Theme} & \textbf{Cont.} & \textbf{Infl.} & \textbf{Role} & \textbf{Inter.} & \textbf{Behav.} & \textbf{Emot.} & \textbf{Cook.} & \textbf{Traff.} & \textbf{Situa.} & \textbf{Acc.} $\uparrow$\\
\midrule
\rowcolor{gray!30}
\multicolumn{16}{c}{\textit{Qwen2.5-VL-3B-Instruct}} \\
\midrule

Batch w/o thinking & 
39.39 & 31.51 & 25.00 & 18.92 & 37.50 & 24.32 & \textbf{43.90} & 20.69 & 29.73 & 16.67 & 42.11 & 38.89 & 30.77 & 10.00 & 27.99 \\

Batch with thinking & 
39.39 & 26.03 & \underline{28.57} & 15.68 & 37.50 & \textbf{40.54} & 17.07 & \textbf{48.28} & 29.73 & 8.33 & \textbf{47.37} & \textbf{61.11} & 38.46 & 20.00 & 28.16 \\

Batch SFT & 
\underline{48.48} & 34.25 & 17.86 & 17.84 & \underline{43.75} & \underline{32.43} & \underline{39.02} & 27.59 & 24.32 & 8.33 & 39.47 & 50.00 & \underline{46.15} & 20.00 & 29.18 \\

Interleaved SFT &
\underline{48.48} & \textbf{38.36} & 25.00 & \textbf{28.65} & 37.50 & 29.73 & 36.59 & \underline{31.03} & \underline{35.14} & \underline{16.67} & 36.84 & \underline{55.56} & \underline{46.15} & \textbf{30.00} & \textbf{33.96} \\

\textbf{TaYS} &
\textbf{51.52} & \underline{36.99} & \textbf{39.29} & \underline{24.86} & \textbf{46.88} & 21.62 & 31.71 & 20.69 & \textbf{37.84} & \textbf{33.33} & \underline{44.74} & 50.00 & \textbf{53.85} & \underline{20.00} & \underline{33.45} \\
\midrule
\rowcolor{gray!30}
\multicolumn{16}{c}{\textit{Qwen2.5-VL-7B-Instruct}} \\
\midrule

Batch w/o thinking & 
54.55 & 32.88 & 28.57 & 14.59 & 37.50 & 27.03 & 41.46 & 34.48 & 29.73 & 16.67 & 36.84 & 52.94 & 46.15 & 10.00 & 28.89 \\

Batch with thinking & 
42.42 & 31.51 & \textbf{42.86} & 23.24 & 37.50 & 29.73 & 29.27 & \textbf{37.93} & 29.73 & \textbf{25.00} & \underline{39.47} & \underline{55.56} & 38.46 & \textbf{30.00} & 31.57 \\

Batch SFT &
48.48 & 26.03 & 32.14 & 21.62 & 43.75 & 32.43 & \underline{43.90} & 31.03 & 27.03 & 8.33 & \underline{39.47} & 38.89 & 46.15 & 20.00 & 30.38 \\

Interleaved SFT &
\underline{57.58} & \textbf{38.36} & \underline{35.71} & \underline{24.86} & \textbf{50.00} & \underline{35.14} & \textbf{48.78} & 24.14 & \underline{37.84} & 16.67 & 36.84 & 44.44 & \underline{46.15} & \underline{20.00} & \underline{34.98} \\

\textbf{TaYS} &
\textbf{63.64} & \underline{35.62} & 28.57 & \textbf{25.95} & \underline{46.88} & \textbf{35.14} & 41.46 & \underline{34.48} & \textbf{51.35} & \underline{16.67} & \textbf{47.37} & \textbf{66.67} & \textbf{46.15} & 10.00 & \textbf{36.86} \\

\bottomrule
\end{tabular}
}
\end{table*}

\paragraph{Parallel KV Cache.}

The core enabler of TaYS's concurrency is a dual-cache system that manages visual and textual states independently. 
We maintain two modality-specific caches: a read-heavy video cache $\mathcal{C}_v$ and a dynamic text cache $\mathcal{C}_r$.

At time step $t$, the incoming frame $F_t$ is processed by the visual encoder and incrementally appended to the video cache:
\[
\mathcal{C}_v^{(t)} 
= 
\mathcal{C}_v^{(t-1)} 
\cup 
\mathrm{Enc}(F_t).
\]
Crucially, this update is non-blocking and occurs asynchronously with respect to the reasoning process.

During the decoding phase, attention is computed over a logical concatenation of the current video cache $\mathcal{C}_v^{(t)}$ and the historical text cache $\mathcal{C}_r^{(t-1)}$. 
We implement this \emph{merge} operation via pointer-level composition rather than physical tensor concatenation, achieving a zero-copy overhead. 
Once the reasoning segment $R_t$ is generated, only the text cache is updated:
\[
\mathcal{C}_r^{(t)} 
= 
\mathcal{C}_r^{(t-1)} 
\cup 
\mathrm{Dec}(R_t),
\]
while the video cache remains immutable during this step. 
The subsequent \emph{split} operation restores the modality-specific cache views, preparing the system for the next cycle.

This architecture establishes a recursive \emph{merge–generate–split} loop. 
While $\mathcal{C}_r$ is engaged in autoregressive token generation, newly arrived frames are independently absorbed into $\mathcal{C}_v$. 
Consequently, the reasoning process is never stalled by visual encoding. 
Compared to the monolithic cache design in batch or interleaved paradigms, TaYS's decoupled cache architecture minimizes critical path latency and enables true parallel streaming, realizing a system where perception and reasoning evolve simultaneously.
\section{Experiments}

\subsection{Experimental Settings}

\paragraph{Video Benchmark.}
We evaluate TaYS on an extended benchmark protocol derived from \textsc{VideoEspresso}, covering temporal, logical, scene, behavioral, and state understanding. 
The benchmark includes tasks such as \emph{Event Dynamics}, \emph{Causal Analysis}, \emph{Theme Analysis}, and realistic applications like \emph{Cooking Process} and \emph{Traffic Analysis}, forming a comprehensive testbed for streaming video reasoning across diverse semantic contexts.

\paragraph{Models and Baselines.}
We implement TaYS on Qwen2.5-VL-3B/7B-Instruct. 
Comparative baselines include: 
(1) \textbf{Batch w/o Thinking}: a supervised model fine-tuned on direct QA pairs; 
(2) \textbf{Batch w/ Thinking}: incorporates frame-referenced intermediate reasoning prompts;\footnote{Detailed CoT inference prompt is provided in Appendix~\ref{sec:prompt_details}.}
(3) \textbf{Batch SFT}: distilled from CoT-annotated data; and 
(4) \textbf{Interleaved SFT}: a streaming variant alternating frame input and reasoning generation without parallel caching. 
This setup isolates the benefits of parallel streaming against conventional batch and sequential interleaving paradigms.

\paragraph{Metrics.}
Evaluation considers both reasoning quality and latency. 
\textbf{Objective performance} requires the semantic similarity of predictions to exceed a threshold and outperform distractors. 
\textbf{Subjective performance} is ranked by GPT-5~\cite{openai2025gpt5} based on logical consistency, factual accuracy, and contextual appropriateness. 
\textbf{Latency} is measured by \emph{TTFT} (time to first token) and \emph{overall delay} (total time for reasoning and response).

\subsection{Results on Benchmark}

\paragraph{Objective Evaluation Results.}
Table~\ref{tab:accuracy} summarizes objective results. 
Explicit CoT prompting enhances base LVLM reasoning, while fine-tuning on temporally aligned trajectories yields further gains by aligning reasoning with visual evidence. 
Streaming-based models outperform all batch baselines significantly. 
Notably, the \emph{Interleaved} model achieves slightly higher accuracy than \emph{TaYS}, suggesting both streaming paradigms effectively capture temporal dependencies. 
However, objective metrics alone may not fully reflect reasoning coherence, necessitating further subjective evaluation.

\begin{figure}[t]
  \centering
  \includegraphics[width=\linewidth]{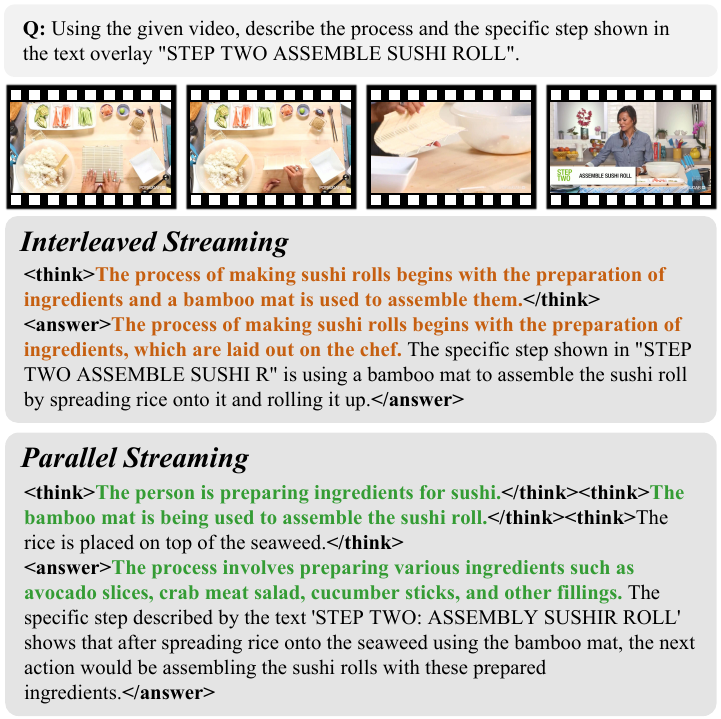}
  \caption{Case study comparing TaYS with the Interleaved paradigm. TaYS produces temporally aligned reasoning, whereas the Interleaved model generates less accurate, fragmented descriptions.}
  \label{fig:Case}
\end{figure}

\paragraph{Subjective Evaluation Results.}
GPT-5 ranked model outputs based on overall quality.\footnote{Detailed subjective evaluation prompt is provided in Appendix~\ref{sec:prompt_details}.} 
TaYS achieved the highest normalized win rate of \textbf{43.7\%}, surpassing Batch (\textbf{31.4\%}) and Interleaved (\textbf{21.7\%}). 
TaYS excels in tasks requiring multi-step temporal reasoning, winning \textbf{61.1\%} of \emph{Cooking Process} samples (vs. 11.1\% for Interleaved) and \textbf{75.0\%} of \emph{Preparation Steps}. 
As illustrated in Figure~\ref{fig:Case}, TaYS aligns reasoning tightly with visual evidence, avoiding the fragmented descriptions produced by the Interleaved model, thereby demonstrating superior temporal grounding in dynamic scenarios.

\begin{table}[t]
  \centering
  \caption{Latency and accuracy comparison across different FPS. TaYS achieves the lowest TTFT and delay, demonstrating superior real-time efficiency.}
  \label{tab:fps_latency}
  \setlength{\tabcolsep}{3.5pt}
  \resizebox{\linewidth}{!}{
  \begin{tabular}{lcccccc}
    \toprule
    \textbf{Method} & \textbf{Metric} & \textbf{FPS=1} & \textbf{FPS=2} & \textbf{FPS=3} & \textbf{FPS=4} & \textbf{FPS=5} \\
    \midrule
    \multirow{3}{*}{Batch} 
      & TTFT$\downarrow$  & 10.36 & 10.48 & 10.62 & 10.77 & 10.93 \\
      & Delay$\downarrow$ & 12.05 & 13.90 & 12.93 & 13.08 & 13.12 \\
      & Acc$\uparrow$     & 28.33 & 29.18 & 31.23 & 30.03 & \underline{31.91} \\
    \midrule
    \multirow{3}{*}{Interleaved}
      & TTFT$\downarrow$  & 0.0303 & 0.0295 & 0.0296 & 0.0301 & 0.0298 \\
      & Delay$\downarrow$ & 12.94 & 14.19 & 16.15 & 18.03 & 20.13 \\
      & Acc$\uparrow$     & \textbf{33.95} & \textbf{33.96} & \underline{33.11} & \underline{31.91} & 30.55 \\
    \midrule
    \multirow{3}{*}{TaYS}
      & TTFT$\downarrow$  & $1\text{e-}6$ & $9.2\text{e-}7$ & $9.3\text{e-}7$ & $1.06\text{e-}6$ & $9.6\text{e-}7$ \\
      & Delay$\downarrow$ & 12.06 & 12.19 & 12.32 & 12.30 & 12.31 \\
      & Acc$\uparrow$     & \underline{31.74} & \underline{33.45} & \textbf{36.01} & \textbf{35.49} & \textbf{34.06} \\
    \bottomrule
  \end{tabular}
  }
\end{table}

\subsection{Real-Time Streaming Reasoning Efficiency}

\begin{figure}[t]
  \centering
  \includegraphics[width=\linewidth]{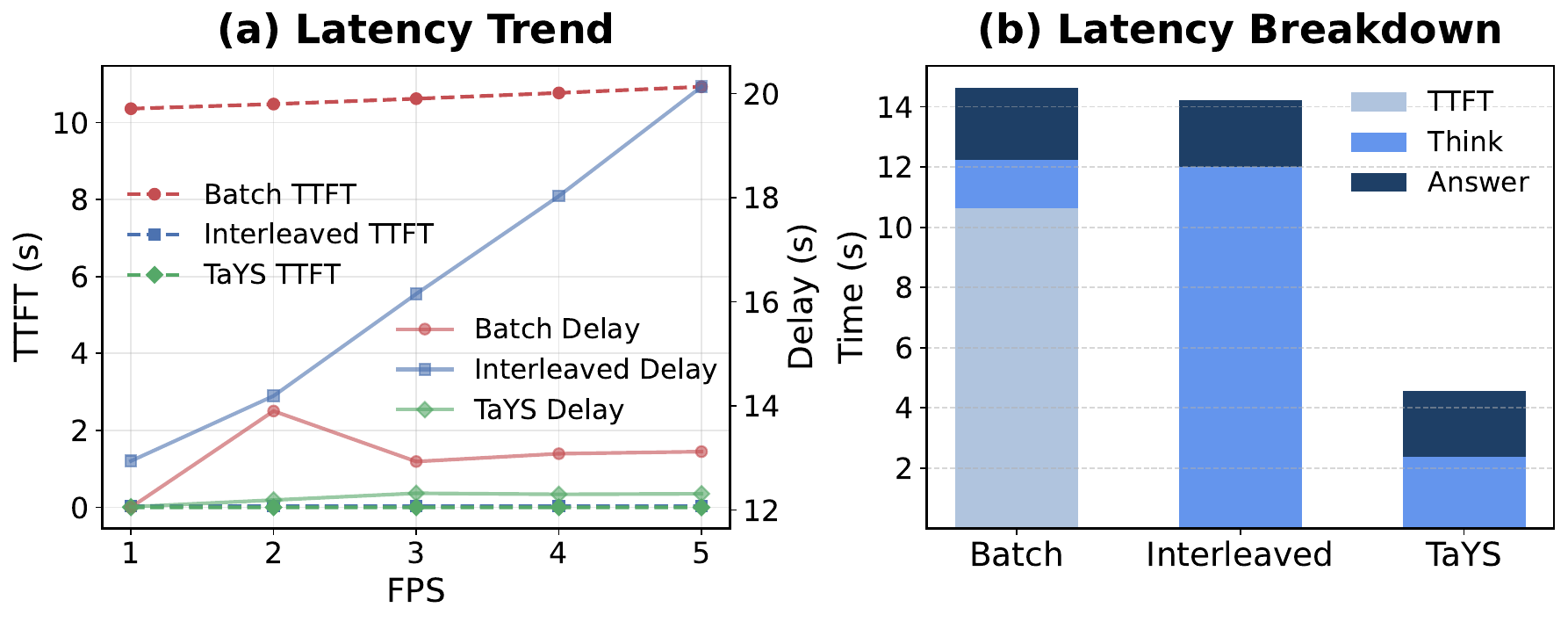}
  \caption{\textbf{(a)} Latency comparison across paradigms. \textbf{(b)} Latency breakdown of TaYS. Parallel KV Cache design enables the lowest TTFT and stable delay.}
  \label{fig:latency_combined}
\end{figure}

We evaluate TaYS in real-time streaming scenarios where frames arrive progressively.
As shown in Table~\ref{tab:fps_latency} and Figure~\ref{fig:latency_combined}(a), the \textbf{Batch} paradigm suffers from a persistent bottleneck ($\sim$10.6s TTFT). 
The \textbf{Interleaved} paradigm responds faster but suffers from cumulative delay growth at higher frame rates due to sequential encode–generate dependencies.

In contrast, \textbf{TaYS} achieves near-zero decoder-level TTFT ($\approx 10^{-6}$s) under the incremental warm-start setting, reflecting minimal decoding latency. 
Crucially, TaYS maintains a stable end-to-end delay of $\sim$12s across all frame rates by parallelizing cache management and reasoning. 
Accuracy scales robustly with frame rate (peaking at 36.0\% for FPS=3), whereas baselines fluctuate. 
Figure~\ref{fig:latency_combined}(b) confirms TaYS's compact latency profile, demonstrating its efficiency and reliability for streaming understanding.

\subsection{Temporal Behavior of Streaming Reasoning}

\begin{figure}
    \centering
    \includegraphics[width=\linewidth]{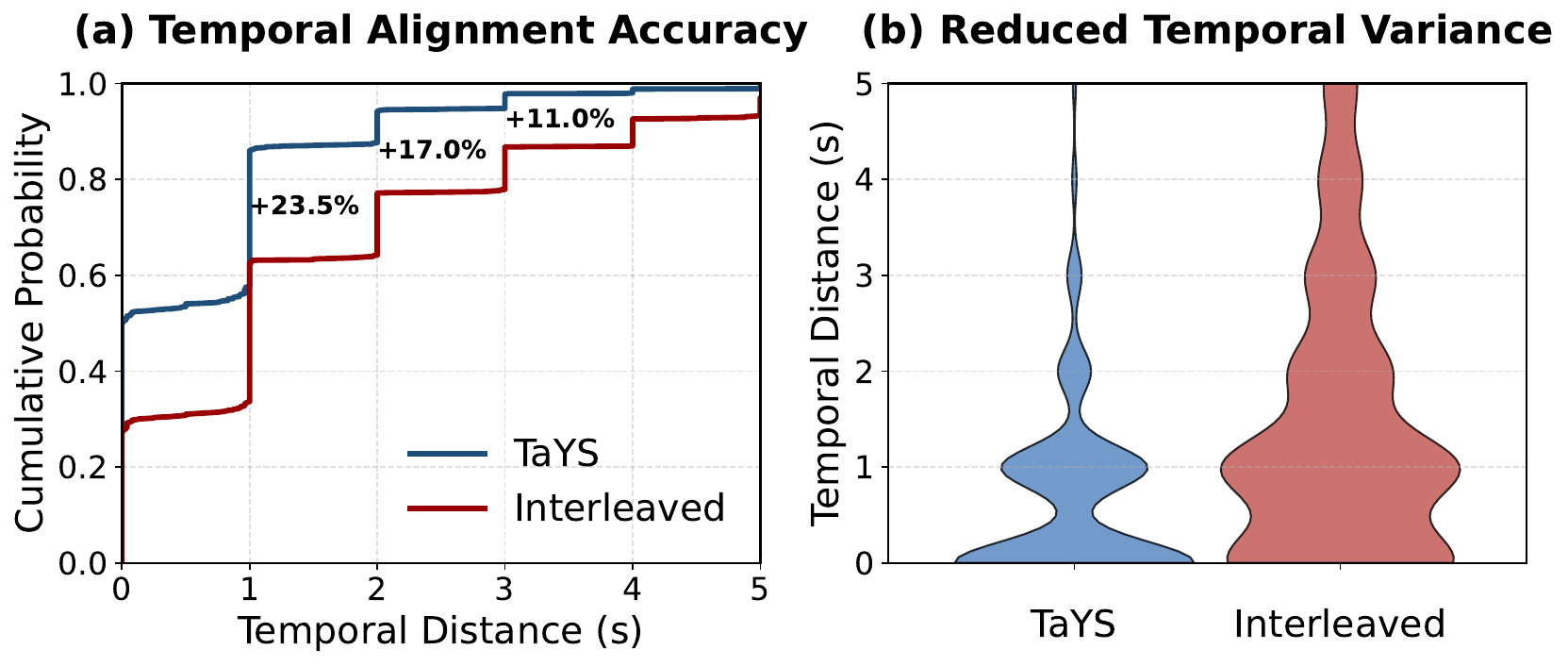}
    \caption{Temporal distance $\Delta t$ distribution. TaYS aligns reasoning more closely with keyframes, achieving higher precision than the interleaved baseline.}
    \label{fig:temporal_alignment}
\end{figure}

\paragraph{Fine-Grained Temporal Alignment.}
We assess whether reasoning is triggered at correct moments by measuring the temporal distance $\Delta t$ between reasoning steps and annotated keyframes. 
Figure~\ref{fig:temporal_alignment} shows TaYS achieves a mean deviation of 0.69s (vs. 1.52s for Interleaved). 
Additionally, 86.0\% of TaYS's reasoning falls within one second of keyframes (vs. 62.4\% for Interleaved). 
The distribution indicates TaYS effectively concentrates reasoning around event boundaries rather than scattering outputs across irrelevant temporal segments, thereby confirming precise temporal grounding and acute event sensitivity.

\begin{figure}
    \centering
    \includegraphics[width=\linewidth]{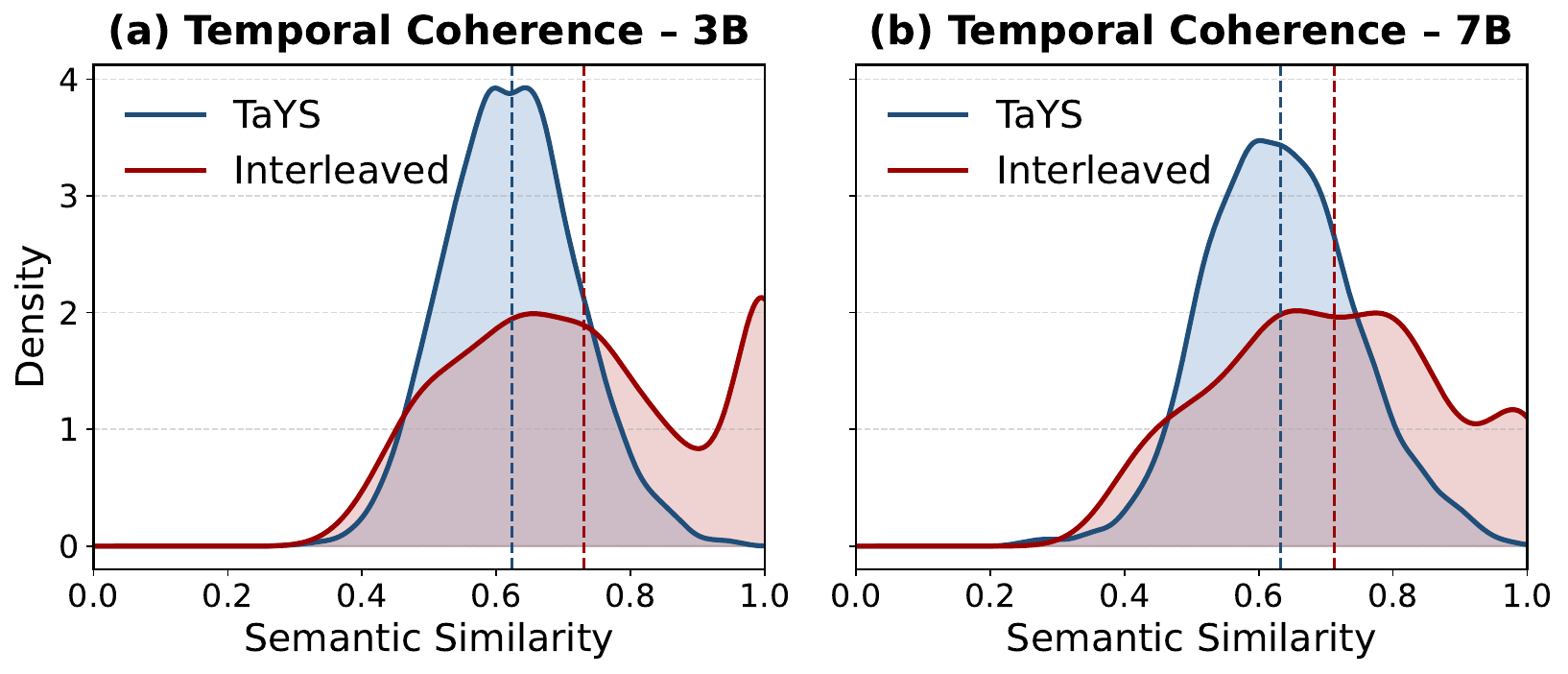}
    \caption{Semantic similarity between consecutive reasoning steps. TaYS maintains a smoother distribution, whereas the interleaved model exhibits repetitive peaks (high similarity), indicating redundancy.}
    \label{fig:temporal_coherence}
\end{figure}

\paragraph{Temporal Coherence of Reasoning.}
We examine semantic continuity between consecutive reasoning outputs (Figure~\ref{fig:temporal_coherence}). 
TaYS exhibits a smooth similarity profile, indicating reasoning evolves with visual changes. 
The suppression of high-similarity spikes suggests effective avoidance of stagnant or looping descriptions, ensuring sustained distinctiveness. 
Conversely, the Interleaved model displays prominent peaks, reflecting redundant and less adaptive reasoning that struggles to assimilate new events. 
These results demonstrate TaYS maintains a coherent, progressive reasoning trajectory aligned with the video's temporal structure.

\section{Conclusion}
Video data naturally arrives as a continuous stream, yet most LVLMs rely on offline batch reasoning, fundamentally misaligned with the sequential nature of real-world visual inputs. 
We introduce the \textbf{streaming thinking paradigm}, enabling models to reason progressively as frames arrive and refine outputs dynamically. 
We instantiate this via \textbf{Think-as-You-See (TaYS)}, integrating streaming Chain-of-Thought, stream-aligned training, and parallel KV-cache architecture. 
Experiments show TaYS reduces latency while enhancing reasoning quality by grounding inferences in immediate visual evidence. 
By decoupling perception from reasoning, our approach resolves the trade-off between responsiveness and depth, allowing models to "think on their feet" without awaiting complete encoding. 
Analyses highlight controllable and temporally grounded reasoning, paving the way for responsive, reliable real-time video understanding. 
This work shifts focus from static analysis to dynamic interaction, laying out a foundation for embodied intelligence and open-world agents.

{
    \small
    \bibliographystyle{ieeenat_fullname}
    \bibliography{main}
}

\clearpage
\setcounter{page}{1}
\maketitlesupplementary

\renewcommand{\thesection}{\Alph{section}}
\setcounter{section}{0}

\section{Details of Streaming CoT Pipeline}
\label{sec:details_cot}

\subsection{CLIP-Guided Frame ID Alignment}

\paragraph{Step 1: Semantic anchoring before resampling.}
Given a video $\mathcal{V}=\{F_t\}_{t=1}^{T}$ with timestamps $\{\tau_t\}_{t=1}^{T}$ and annotated keyframe captions $\mathcal{C}=\{c_k\}_{k=1}^{K}$, we first compute CLIP embeddings for all frames and captions:
\[
\boldsymbol{f}_t = \mathrm{Enc}_{\text{CLIP}}^{\text{img}}(F_t), 
\qquad 
\boldsymbol{g}_k = \mathrm{Enc}_{\text{CLIP}}^{\text{text}}(c_k).
\]
We utilize cosine similarity throughout the alignment process:
\[
\mathrm{sim}(\boldsymbol{a},\boldsymbol{b}) \;=\; 
\frac{\boldsymbol{a}^\top \boldsymbol{b}}{\|\boldsymbol{a}\|\,\|\boldsymbol{b}\|}.
\]
For each keyframe caption $c_k$, we identify its most similar frame index:
\[
t_k^\star \;=\; \arg\max_{\,t\in\{1,\dots,T\}} \;\mathrm{sim}(\boldsymbol{f}_t,\boldsymbol{g}_k),
\]
recording the anchor timestamp $\widehat{\tau}_k = \tau_{t_k^\star}$. These anchors serve as semantic locks preserved during subsequent resampling.

\paragraph{Step 2: Timestamp-based resampling at 2 FPS with anchor preservation.}
Let the target sampling interval be $\Delta=0.5$\,s (2 FPS) and the target grid be $\{\tau'_{t'}\}_{t'=1}^{T'}$ with $\tau'_{t'}=(t'-1)\Delta$. For each target timestamp $\tau'_{t'}$, we select the frame $F_{t'}$ as:
\[
F_{t'} \;=\;
\begin{cases}
F_{t_k^\star}, ~~~~\text{if } \tau'_{t'} \in [\widehat{\tau}_k-\epsilon,\;\widehat{\tau}_k+\epsilon] \text{ for some } k,\\[2pt]
\displaystyle \arg\min_{F_t}\;|\tau_t - \tau'_{t'}|, ~~~~\text{otherwise},
\end{cases}
\]
where $\epsilon=0.1$\,s is a tolerance window ensuring every semantic anchor $\widehat{\tau}_k$ snaps to the nearest sampling point. Post-selection, frame indices are renormalized, and clips are truncated to the maximum input duration (30\,s).

\subsection{Quality Assurance and Temporal Filtering}

To ensure generated frame-level trajectories are temporally grounded and semantically reliable, we apply a three-stage filtering process (Algorithm~\ref{alg:quality_filtering}). First, we identify question-relevant keyframes via embedding similarity. Second, we prune temporally adjacent captions with redundant semantics to preserve distinct perceptual events. Finally, we format the supervision sequence by assigning \texttt{</EOT>} to selected keyframes and \texttt{<SKIP>} to others. This yields a temporally sparse but well-aligned target stream, guiding the model to reason only at meaningful moments.

\begin{algorithm}[t]
\caption{Quality Assurance and Temporal Filtering}
\label{alg:quality_filtering}
\begin{algorithmic}[1]

\Require Question $Q_t$, keyframe captions $\{c_k\}$
\Require Thresholds $\tau_q = 0.7$, $\tau_{\mathrm{adj}} = 0.9$

\Statex \textbf{Step 1: Question–caption relevance screening}
\For{each caption $c_k$}
    \State $s_k \gets \mathrm{sim}(e(Q_t), e(c_k))$
\EndFor
    \State $\mathcal{K}_t \gets \{k \mid s_k \ge \tau_q\}$

\Statex \textbf{Step 2: Anti-redundancy temporal de-duplication}
\State Sort $\mathcal{K}_t$ by time
\State $\mathcal{K}^\star_t \gets [\ ]$
\For{each $k$ in $\mathcal{K}_t$}
    \If{$\mathcal{K}^\star_t$ is empty}
        \State Append $k$ to $\mathcal{K}^\star_t$
    \Else
        \State Let $j$ be last element in $\mathcal{K}^\star_t$
        \State $s_{j,k} \gets \mathrm{sim}(e(c_j), e(c_k))$
        \If{$s_{j,k} < \tau_{\mathrm{adj}}$}
            \State Append $k$ to $\mathcal{K}^\star_t$
        \EndIf
    \EndIf
\EndFor

\Statex \textbf{Step 3: Formatting supervision targets}
\For{each sampled frame index $t'$}
    \If{$t' \in \mathcal{K}^\star_t$}
        \State Emit $[R_{t'}]\ </EOT>$
    \Else
        \State Emit \texttt{<SKIP>}
    \EndIf
\EndFor

\end{algorithmic}
\end{algorithm}

\subsection{Practical Notes}
\begin{itemize}
  \item \textbf{Embedding normalization.} All embeddings are $\ell_2$-normalized prior to similarity computation to stabilize thresholds.
  \item \textbf{Batching.} Frame and caption embeddings are computed in batches to mitigate I/O latency for long videos.
  \item \textbf{Hyperparameters.} Default values are $\Delta=0.5$\,s, $\epsilon=0.1$\,s, $\tau_q=0.7$, and $\tau_{\mathrm{adj}}=0.9$, balancing temporal precision with retention of key semantic content.
\end{itemize}

\subsection{Details of Dataset}

\begin{figure}[h]
    \centering
    \includegraphics[width=\linewidth]{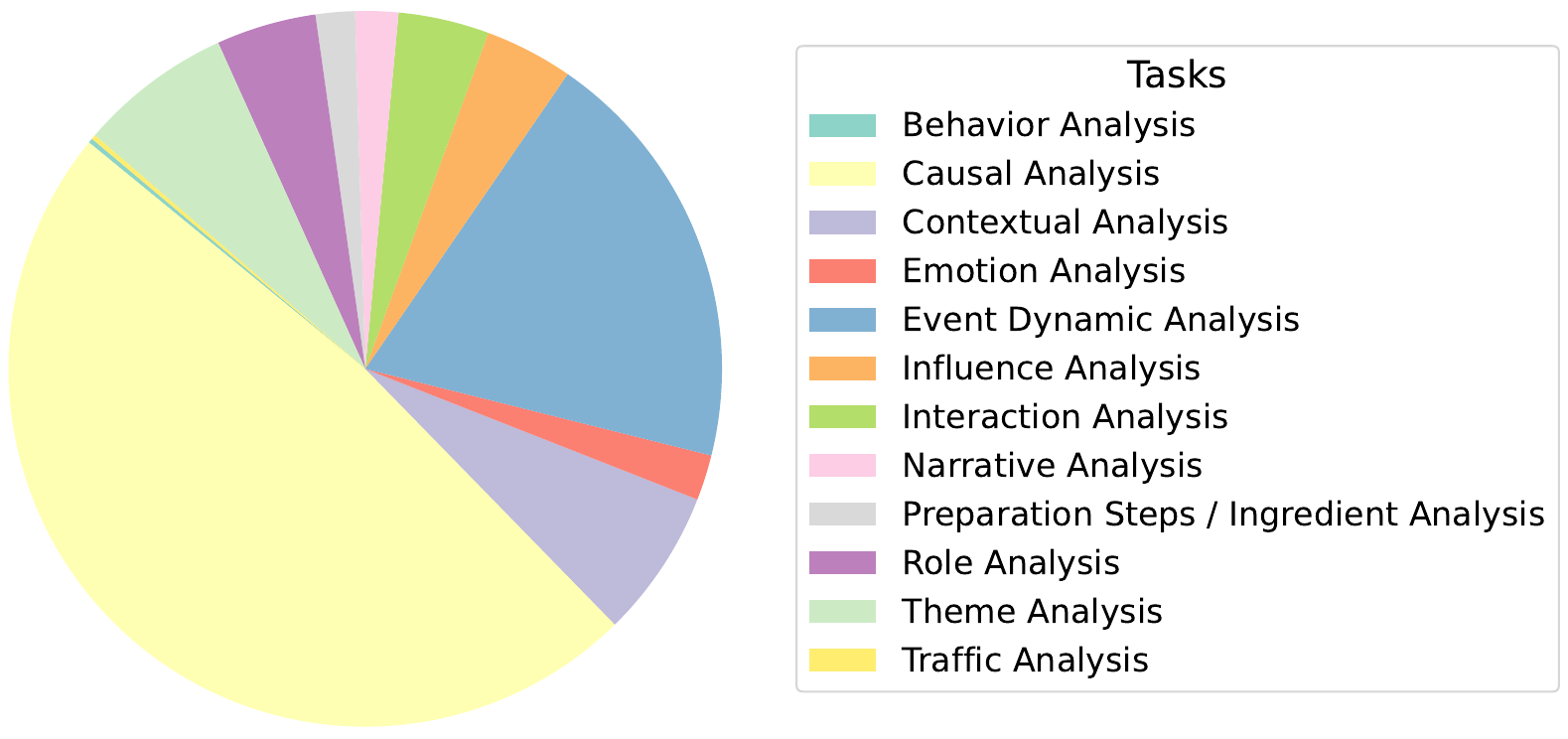}
    \caption{Task distribution in the dataset.}
    \label{fig:dataset}
\end{figure}

The dataset spans 12 video reasoning tasks covering fine-grained event interpretation and high-level semantic understanding. As shown in Figure~\ref{fig:dataset} and Table~\ref{tab:dataset}, the task distribution is long-tailed: \emph{Causal Analysis} and \emph{Event Dynamic Analysis} dominate, while \emph{Ingredient Analysis} and \emph{Behavior Analysis} are less frequent. This reflects the natural prevalence of reasoning behaviors in real-world video content while ensuring broad coverage for multi-step reasoning evaluation.

\begin{figure}[t]
    \centering
    \includegraphics[width=\linewidth]{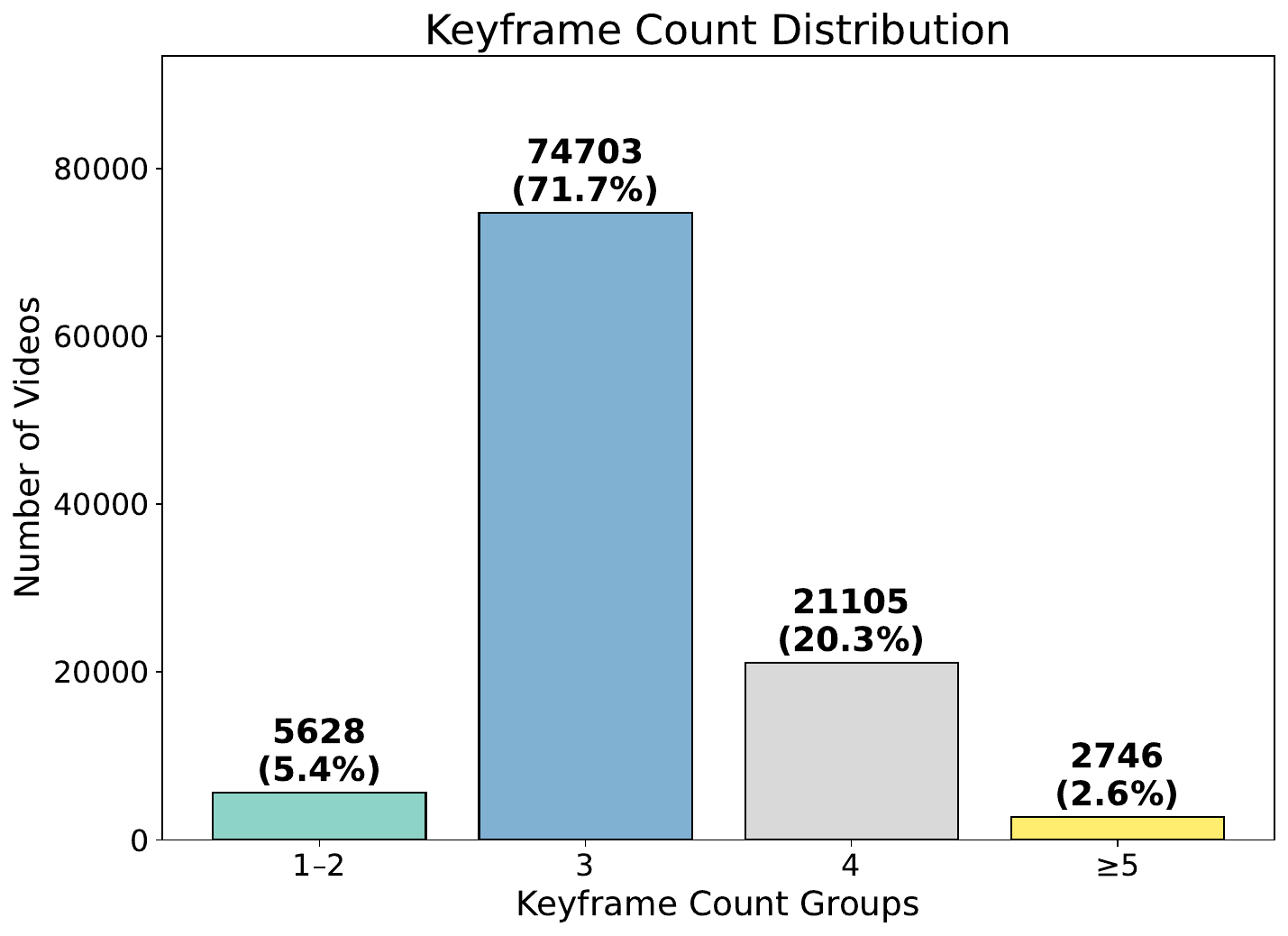}
    \caption{Distribution of keyframe counts per sample.}
    \label{fig:keyframe_distribution}
\end{figure}

Temporal structure also varies significantly. Figure~\ref{fig:keyframe_distribution} illustrates the distribution of keyframe counts, revealing a wide spectrum of temporal sparsity. Some videos contain sparse salient moments, while others feature dense, extended event sequences. This variability is critical for evaluating streaming reasoning, requiring models to adapt to varying event frequencies and accurately identify meaningful visual changes.

\begin{table}[h]
\centering
\small
\setlength{\tabcolsep}{4pt}
\caption{Distribution of task categories in training and test sets.}
\label{tab:dataset}
\begin{tabular}{m{3.8cm} >{\centering\arraybackslash}m{1.8cm} >{\centering\arraybackslash}m{1.8cm}}
\toprule
\textbf{Task} & \textbf{Train Set} & \textbf{Test Set} \\
\midrule
Causal Analysis & 52,566 & 208 \\
Event Dynamic Analysis & 18,675 & 82 \\
Preparation Steps / Ingredient Analysis & 2,252 & 74 \\
Theme Analysis & 6,206 & 33 \\
Interaction Analysis & 4,208 & 38 \\
Influence Analysis & 4,406 & 45 \\
Role Analysis & 4,843 & 31 \\
Emotion Analysis & 1,999 & 39 \\
Narrative Analysis & 1,755 & 35 \\
Contextual Analysis & 6,827 & 38 \\
Behavior Analysis & 227 & 12 \\
Traffic Analysis & 218 & 14 \\
\bottomrule
\end{tabular}
\end{table}

\section{Prompt Details}
\label{sec:prompt_details}

\begin{figure}[h]
    \centering
    \includegraphics[width=\linewidth]{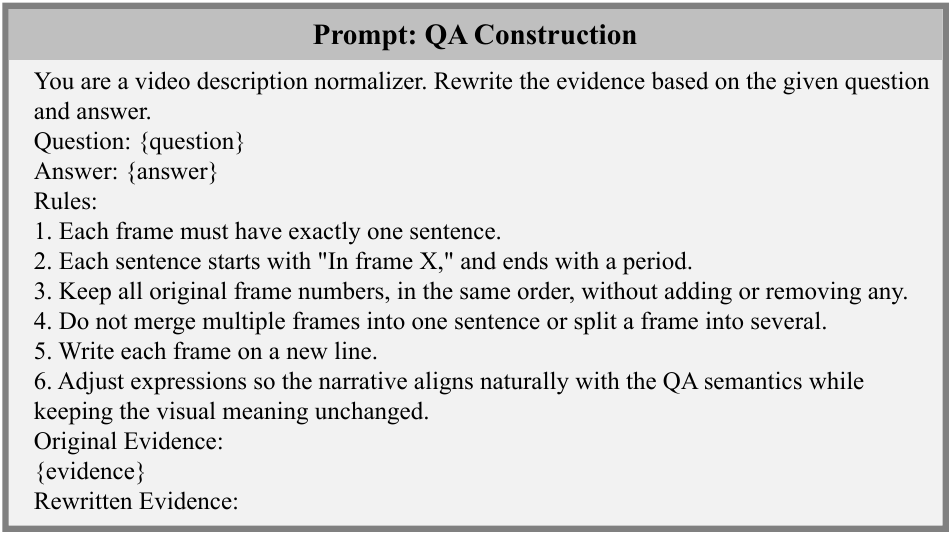}
    \caption{Prompt template for QA construction.}
    \label{fig:Prompt_1}
\end{figure}

\begin{figure}[h]
    \centering
    \includegraphics[width=\linewidth]{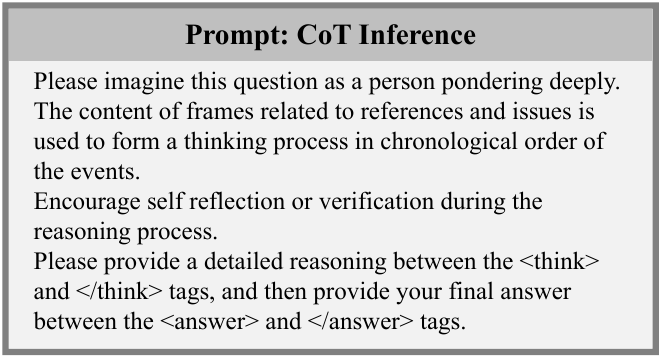}
    \caption{Prompt template for CoT inference.}
    \label{fig:Prompt_2}
\end{figure}

\begin{figure}[h]
    \centering
    \includegraphics[width=\linewidth]{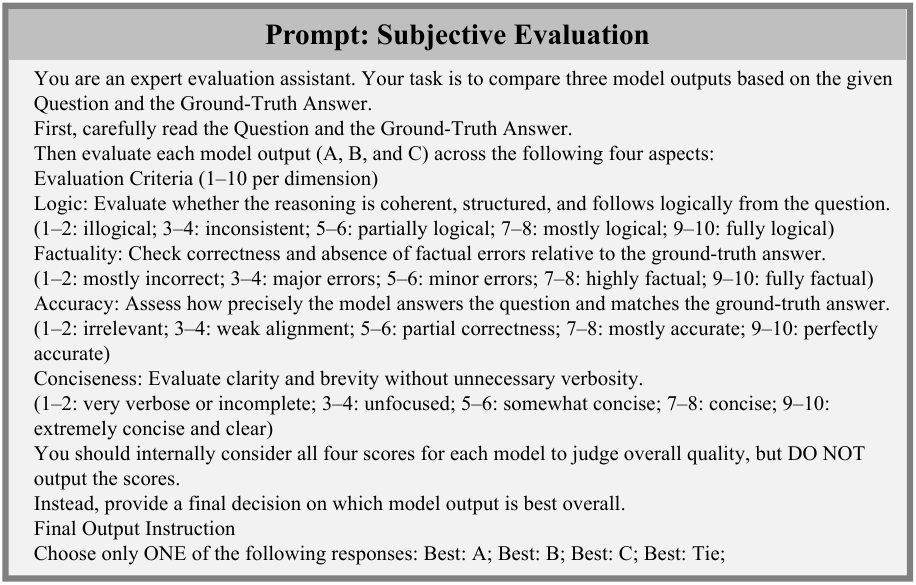}
    \caption{Prompt template for subjective evaluation.}
    \label{fig:Prompt_3}
\end{figure}

We present the complete prompts used in our pipeline, including QA construction (Figure~\ref{fig:Prompt_1}), CoT inference (Figure~\ref{fig:Prompt_2}), and subjective evaluation (Figure~\ref{fig:Prompt_3}).

\section{Training Details}
\label{sec:training_details}

We train TaYS using a streaming-aware decoder-only objective, where visual and reasoning tokens are interleaved with causal masking. Optimization employs AdamW with cosine decay, mixed-precision (bfloat16), gradient accumulation, activation checkpointing, and DeepSpeed ZeRO-3 for memory efficiency. The vision encoder remains frozen, while the multimodal projector and LLM backbone are fine-tuned. We regulate video token length via pixel-based constraints and train for two epochs with an effective sequence length of 8192 tokens.

\begin{table}[h]
\centering
\small
\setlength{\tabcolsep}{6pt}
\caption{Training hyperparameters for TaYS.}
\begin{tabular}{lc}
\toprule
\textbf{Config} & \textbf{Value} \\
\midrule
input resolution            & variable (pixel-constrained) \\
max token length            & 8192 \\
vision encoder              & frozen \\
trainable modules           & LLM + MLP projector \\
precision                   & bfloat16 \\
optimizer                   & AdamW \\
learning rate               & $2\times10^{-5}$ \\
lr schedule                 & cosine decay \\
warmup ratio                & 0.03 \\
batch size                  & 1 (grad accum = 16) \\
epochs                      & 2 \\
gradient clipping           & 1.0 \\
gradient checkpointing      & enabled \\
distributed training        & torchrun + ZeRO-3 \\
max video frames            & 60 \\
video token budget          & 24K tokens (pixel-based) \\
\bottomrule
\end{tabular}
\end{table}

\section{Evaluation Details}
\label{sec:eval_details}

\paragraph{Construction of Test Set.}
Following the VideoEspresso protocol, we construct the test set with three distractor options per question. Distractors are designed to match the correct answer in contextual relevance and linguistic form while containing explicit factual inaccuracies, ensuring a discriminative evaluation. We apply the same answer-rewriting procedure as in training to maintain consistency.

\begin{algorithm}[h]
\caption{Two-Stage Objective Evaluation}
\begin{algorithmic}[1]
\Require Prediction $\tilde{y}$, reference answer $y^\star$, 
         options $\mathcal{O} = \{o_1,o_2,o_3,o_4\}$, 
         correct option $o^\star \in \mathcal{O}$, 
         similarity function $\mathrm{sim}$, threshold $\tau$
\State $s_{\mathrm{ref}} \gets \mathrm{sim}(\tilde{y}, y^\star)$
\If{$s_{\mathrm{ref}} < \tau$}
    \State \Return \textsc{Incorrect}
\EndIf
\For{each $o_j \in \mathcal{O}$}
    \State $s_j \gets \mathrm{sim}(\tilde{y}, o_j)$
\EndFor
\State $s_{\mathrm{opt}} \gets \mathrm{sim}(\tilde{y}, o^\star)$
\State $s_{\max}^{\mathrm{neg}} \gets \max\{ s_j : o_j \in \mathcal{O},\, o_j \neq o^\star \}$
\If{$s_{\mathrm{opt}} \ge \tau$ \textbf{and} $s_{\mathrm{opt}} > s_{\max}^{\mathrm{neg}}$}
    \State \Return \textsc{Correct}
\Else
    \State \Return \textsc{Incorrect}
\EndIf
\end{algorithmic}
\end{algorithm}

\paragraph{Objective Evaluation Protocol.}
For each sample, we evaluate a free-form prediction $\tilde{y}$ against a reference answer $y^\star$ and multiple-choice options $\mathcal{O} = \{o_1,o_2,o_3,o_4\}$, where $o^\star$ is the correct option. We use a semantic similarity function $\mathrm{sim}(\cdot,\cdot)$ with a threshold $\tau = 0.8$.

\textbf{Stage~1: Reference similarity.} We first compute $s_{\mathrm{ref}} = \mathrm{sim}(\tilde{y}, y^\star)$. If $s_{\mathrm{ref}} < \tau$, the prediction is deemed incorrect.

\textbf{Stage~2: Option discrimination.} We compute similarities $s_j = \mathrm{sim}(\tilde{y}, o_j)$ for all options. Let $s_{\mathrm{opt}} = \mathrm{sim}(\tilde{y}, o^\star)$ and $s_{\max}^{\mathrm{neg}} = \max_{o_j \neq o^\star} s_j$. A prediction is correct only if:
\[
s_{\mathrm{ref}} \ge \tau, \quad
s_{\mathrm{opt}} \ge \tau, \quad
\text{and} \quad
s_{\mathrm{opt}} > s_{\max}^{\mathrm{neg}}.
\]

\paragraph{Latency Evaluation Protocol.}
We quantify real-time performance using two metrics: 
\textbf{(1) Time to First Token (TTFT)}, measuring the interval between the arrival of the first frame and the emission of the first token; 
\textbf{(2) Overall Delay}, measuring the total time to complete reasoning and produce the final answer. 
All inferences run on identical hardware with token-level timing resolution to ensure fair comparison.

\end{document}